\documentclass[conference]{IEEEtran}
\usepackage[backend=biber, maxnames=1, style=numeric-comp, isbn=false, url=false, doi=false, sorting=none]{biblatex}
\usepackage[pdftex]{graphicx}
\usepackage[cmex10]{amsmath}

\ifCLASSOPTIONcompsoc
  \usepackage[caption=false,font=normalsize,labelfont=sf,textfont=sf]{subfig}
\else
  \usepackage[caption=false,font=footnotesize]{subfig}
\fi

\usepackage{url}

\renewcommand\IEEEkeywordsname{Keywords}

\begin{document}

\title{Assessing aesthetics of generated abstract images using correlation structure}

\author{\IEEEauthorblockN{Sina Khajehabdollahi}
\IEEEauthorblockA{Eberhard Karls University of\\
T\"ubingen, Germany\\
Email: sina.abdollahi@gmail.com}
\and
\IEEEauthorblockN{ Georg Martius}
\IEEEauthorblockA{Max Planck Institute for Intelligent Systems\\
T\"ubingen, Germany\\
Email: georg.martius@tuebingen.mpg.de}
\and
\IEEEauthorblockN{Anna Levina }
\IEEEauthorblockA{Eberhard Karls University of\\
T\"ubingen, Germany\\
Email: anna.levina@uni-tuebingen.de }
}

\maketitle

\begin{abstract}
Can we generate abstract aesthetic images without bias from natural or human selected image corpi?
Are aesthetic images singled out in their correlation functions?
In this paper we give answers to these and more questions.
We generate images using compositional pattern-producing networks with random weights and varying architecture. We demonstrate that even with the randomly selected weights the correlation functions remain largely determined by the network architecture.

In a controlled experiment, human subjects picked aesthetic images out of a large dataset of all generated images.
Statistical analysis reveals that the correlation function is indeed different for aesthetic images.

\end{abstract}


\IEEEpeerreviewmaketitle

\section{Introduction}

 We all marvel at the aesthetic beauty of natural landscapes, the statistics of these images shapes our understanding of beauty. To which extent different statistical measures of the image affect the perception of beauty remains a big question. Photographs of the real world or figurative art mixes the interpretation of the objects and the narrative of the image with one's aesthetic perception. Investigating how generated abstract images are perceived by human observers allows one to uncover the underlying principles.

Natural images follow universal statistical characteristics manifested by specific scaling in the power-spectrum and in the two-point correlations~\cite{ruderman1994statistics}.
 However, even after two-point correlations are removed, natural images remain recognizable due to higher order relations~\cite{olshausen1996emergence}. Still, two-point correlations affect our perception both in natural and artificial images.
Here we aim to understand, on the one side, to which extent such correlations relate to aesthetic attractiveness of images and on the other side, how specific features of artificial neural networks can be used to generate specific statistics of correlations.

The automatic generation of images is typically designed to mimic real scenes, e.g.~by attempting
to represent the probability density of a large set of images using deep networks and variational inference (Variational Autoencoder~\cite{Kingma2013AutoEncodingVB} or by adversarial training (Generative Adversarial Nets~\cite{Goodfellow2014:GANS}).
However, for answering our questions, images generated in this way are not suitable, as they do not allow  to disentangle the effects of correlation statistics and concrete object interpretations.
Ideally, we need a way to generate abstract images without anchoring them on predefined examples.

Compositional Pattern-Producing Networks~\cite{stanley2007compositional} (CPPNs) are capable of generating a large variety of images without being trained on a particular set of scenes.
Images generated by specially trained/evolved CPPNs~\cite{secretan2008picbreeder} were shown to be identified as natural objects both by humans as well as deep neural networks~\cite{nguyen2015deep} albeit being of abstract nature. On the other side, without training or evolutionary selection CPPNs simply transform and decorrelate highly correlated inputs.
Interestingly, these neural network with random weights create structured outputs that vary significantly among
 architectures and hyper-parameters.

We investigate, how constraining the layout of CPPNs in terms of numbers of layers and number of neurons per layer shapes the output correlations. We demonstrate that the ensemble of such networks can generate the breadth of correlation structures similar to the ones observed over different types of natural images. We assess the attractiveness of the generated images to human subjects and evaluate how it is connected with the statistics of two-point correlations.

\section{Methods}

\begin{figure}
    \centering
    \includegraphics[width=0.96\linewidth]{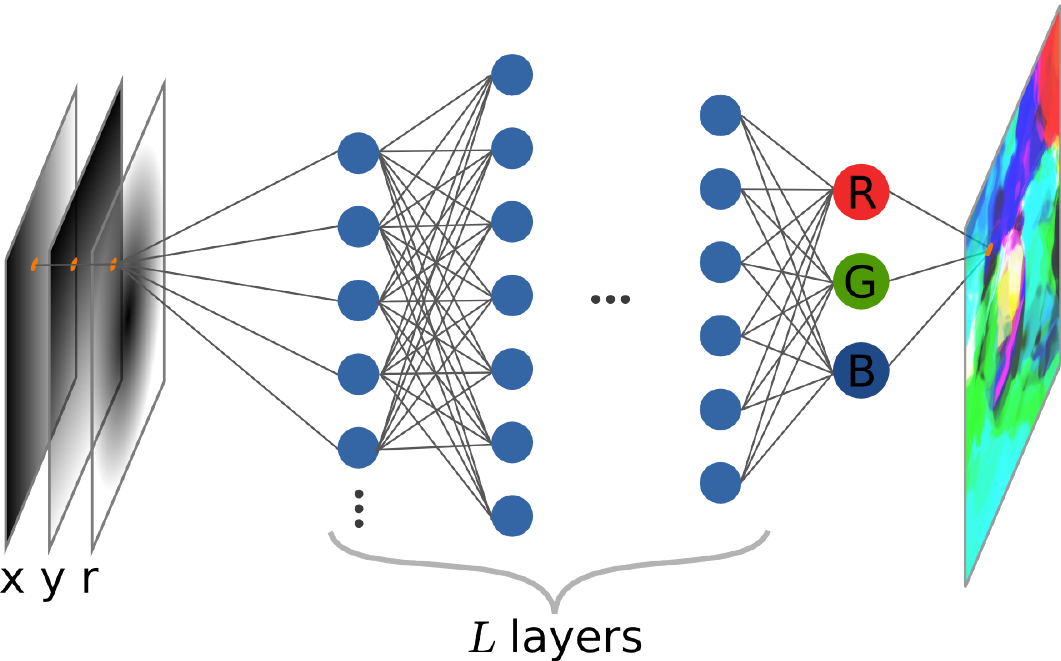}
    \caption{Illustration of the CPPN architecture. The network processes each pixel independently, here parametrized by the coordinates $x,y$ and radius $r$. We vary the number of layers, $L$, and the total number of neurons, $N$ (all blue), and the relative layer sizes.}
    \label{fig:arch}
\end{figure}

\subsection{Generating images with CPPNs}
We use Compositional Pattern-Producing Networks~\cite{stanley2007compositional} (CPPNs) with random weights to generate images of size 512 pixels by 512 pixels. The architecture is illustrated in Fig.~\ref{fig:arch}.
To explore a variety of image styles, the architectures of the CPPNs are varied.
To allow for a controlled generation of architectures we parametrize different aspects with
 five hyper-parameters.
$L \in \{3,5,10\}$ is the parameter defining the total number of layers in the architecture and $N \in \{100, 250, 500\}$ is the total number of neurons, see also Fig.~\ref{fig:arch}.

An initial trial-and-error search suggested that interesting results can be obtained
 by varying number of neurons in each layer.
Thus, the following equation expresses the number of neurons $n(l)$ in layer $l$ as:
\begin{equation}\label{eq:net_arch}
    n(l) = C_N N e^{\mu l / L} \left ( \alpha + \sin (-\omega l)  \right).
\end{equation}
$\mu \in \{-1, -0.5, -0.1, 0, 0.1, 0.5, 1\}$ is the decay rate parameter that specifies how the number of neurons decreases (or increases), $\omega \in \{-2, -1, -0.5, 0, 0.5, 1, 2\}$ is the frequency parameter that allows us to introduce bottlenecks to the architecture, $\alpha \in \{2, 5\}$ is a parameter that controls the strength of the oscillation terms, and $C_N$ is a normalizing factor that ensures the total number of neurons in the architecture is as close as possible to $N$ while making sure that no layers have less than 2 neurons (which often results in a simple, solid coloured image).

\subsubsection{Network details}
The layers are fully connected and each neuron also has a bias weight.
Activation functions are hyperbolic tangents with cubed inputs ($f(z) = \tanh(z^3)$) in the hidden layers and a sigmoid for the output layer (for red, green, blue colour channels).
Once a network architecture is defined, all weights are sampled from a normal distribution with mean 0 and standard deviation 1.
An image is generated by feeding in each pixel independently in form of $(x, y, r)$
 where $x\in [-1,1]$, $y\in [-1,1]$ are the 2D coordinates (with $(0,0)$ at the center) and $r = \sqrt{x^2 + y^2}$ is the radius.

Due to the multi-layer structure and random weights the network transforms the simple inputs and creates intricate images.
To give an intuition, a particular network might give larger weights to the $r$ component initially, giving rise to images whose structures are more circular and symmetric. Further layers act to distort the image. Stronger weights give rise to noisier shapes and colours and smaller weights to simple colours and shapes which tend to be smoother.

\subsubsection{Generated image database}
We generated a large dataset of 35280 images, 40 images per 882 unique architectures.
For each architecture configuration, 40 images are generated by changing the random number generator (RNG) seed. These seeds are then re-used for different architecture configurations to allow for easier comparison.

\subsection{Image analysis}

In the following we detail the statistical tools and methods used to analyze
both generated images as well as natural images.
The main measure of choice is the spatial correlation function which we efficiently calculate using a convolution method.

In order to answer our research questions, we asked 45 human participants to select aesthetic images as detailed in Sec.~\ref{sec:aest:sel}.
The correlation functions of these aesthetic images were then compared to both the full dataset and to more restricted architecture subsets.
Furthermore, a set of natural images (photographs of nature, cities, animals, landscapes and the like) are compared to the CPPN-generated dataset based in their correlation function.
To compute whether a certain set of images has a significantly different correlation function
 than another set, we are using a Welch's test (similar to a t-test but for samples of varying size and variance).

\subsubsection{Correlation function}

To calculate the correlation function, the Pearson correlation coefficient Eq.~\ref{eq:pearsonr} of each image is calculated as a function of pixel-wise distance using only the luminosity information (not taking into account colours). For example, for a distance of 1 pixel, all possible pixel pairs that have a distance of 1 would be put into the two vectors $X, Y$ where the Pearson correlation coefficient is calculated as:

\begin{align}
    \begin{split}\label{eq:pearsonr}
        \rho_{X, Y} &= \frac{\mathrm{E}\left[\left(X-\mu_{X}\right)\left(Y-\mu_{Y}\right)\right]}{\sigma_{X} \sigma_{Y}}
    \end{split}\\
    \begin{split}\label{eq:pearsonrexp}
        &= \frac{\mathrm{E}[XY]-\mathrm{E}[X] \mathrm{E}[Y]}{\sqrt{\mathrm{E}\left[X^{2}\right]-[\mathrm{E}[X]]^{2}} \sqrt{\mathrm{E}\left[Y^{2}\right]-[\mathrm{E}[Y]]^{2}}}
    \end{split}
\end{align}

where $\mathrm{E}[X]$ denotes the expectation of random variable $X$, $\mu_{X} = \mathrm{E}[X]$ and $\sigma_{X}$ are the mean and the standard deviation of $X$.
Note that the larger the distance the fewer pairs exist.
Conceptually, we will use the correlation coefficients for a given distance, yielding
 a correlation function: distance vs.~correlation coefficient for each image.
In the remaining of this section we elaborate on how to make these computations efficient. Understanding this is not essential and might be skipped.

FFT-convolution methods can be employed to quickly calculate the terms in Eq.~\ref{eq:pearsonrexp} without the need to extract pairs explicitly.
In the simplest case, one can convolve an image with its horizontally and vertically flipped version (using python numpy syntax): {\small \verb|convolve(image, image[::-1, ::-1])|}, to obtain a non-normalized version of a correlation coefficient. However, to obtain the Pearson correlation coefficient, normalization with respect to the number of pixels involved at each distance/angle, sample means and sample variances must also be computed as in the terms in Eq.~\ref{eq:pearsonrexp}. The number of pixels contributing to the correlation coefficient for a distance of $x$ and $y$ pixels horizontally and vertically is then given by $N(x,y) = $ \verb|convolve(I, I)| where \verb|I| is a matrix of ones the same shape as the image. The sample means are defined as $\mu(x,y) = $ \verb|convolve(image, I)/N(x,y)| where we also compute the flipped version and denote it as $\mu(-x, -y)$. If we let $\vec{a}=(x,y)$ then we can write the correlation coefficient as:

\begin{align}\label{eq:cc-convolve}
    \rho(\vec{a})=\frac{\sum_{\vec{p}=(1,1)}^{(N,N)-\vec{a}} i(\vec{p}+\vec{a}) i(\vec{p})+N(\vec{a})[-\mu(\vec{a}) \mu(-\vec{a})]}{N(\vec{a}) \sigma(\vec{a}) \sigma(-\vec{a})},
\end{align}
where $i(\vec{p})$ is luminescence of the image at position $\vec{p}$. All terms can be calculated using convolution operations. The output of this method gives a matrix that is twice the resolution of the input image, however due to the symmetry in translations of $a$ and $-a$ we can just take the upper diagonal of this matrix, giving us $\left ((1024 \times 1024) - 1024 \right) / 2 = 523776$ correlation coefficients per image.
The correlation coefficients obtained now have both a distance and angle to which they correspond to. Typically we average out the angle information yielding a single curve: distance vs.~correlation coefficient.

\subsubsection{Aesthetic selection}\label{sec:aest:sel}

To identify `aesthetic' images in the CPPN dataset we designed the following experiment.
We consider subsets of 200 randomly sampled images, two per participant, one coloured and one transformed into grayscale.
We recruited 45 participants, each shown a unique pair of of set. Each participant had a maximum time of 10 minutes to go through one set of images (20 minutes total for both sets) where they are instructed to tag any image they deem aesthetic/beautiful/attractive (we gave no strict criteria to the participants inviting them to use the most appropriate synonym) using the ``XnView'' program~\cite{XnView}. No other criteria was prescribed. Participants could scroll through the set and edit the tags as much as they wanted up to the maximal time, or finish rating earlier.

1403 grayscale images were tagged as aesthetic (from the 9000 grayscale images presented), and 1314 colour images were tagged as aesthetic (again from the 9000 colour images presented).
A selection of tagged images is presented in Fig.~\ref{fig:many_pics}.

\subsubsection{Natural images}

To compare to natural images, we have chosen the McGill Calibrated Colour Image Database \cite{olmos2004biologically} and used the sets: Animals, Flowers, Foliage, LandWater, ManMade, amd Snow. These images were scaled down so that their smallest dimension was 512 pixels, and center-cropped so that their final resolution was exactly 512 by 512 pixels.

\subsubsection{Welch's test}
To answer whether the correlation functions for two different image sets are significatly different
 a one-sided Welch's test was used.
 For presentation purposes we averaged correlation functions from images generated by the same architecture (see Fig.~\ref{fig:correlation}).

 In the case of the aesthetic set we pool together all the 1403 (1314) of grayscale (coloured) images tagged by the participants. For the natural images, each category has anywhere from 150 up to 1112 images. The correlation functions are then binned into 512 distance bins such that the different orientations and discrete nature of the square images are coalesced. We use a Bonferroni correction for multiple comparison: the threshold of significance is divided by a factor proportional to the number of bins within our scope of interest. When comparing the full datasets these are the bins up to a distance of 300 and when comparing the architecture specific datasets these are the bins up to a distance of 100.

\section{Results}

\begin{figure}
    \centering
    \includegraphics[width=0.96\linewidth]{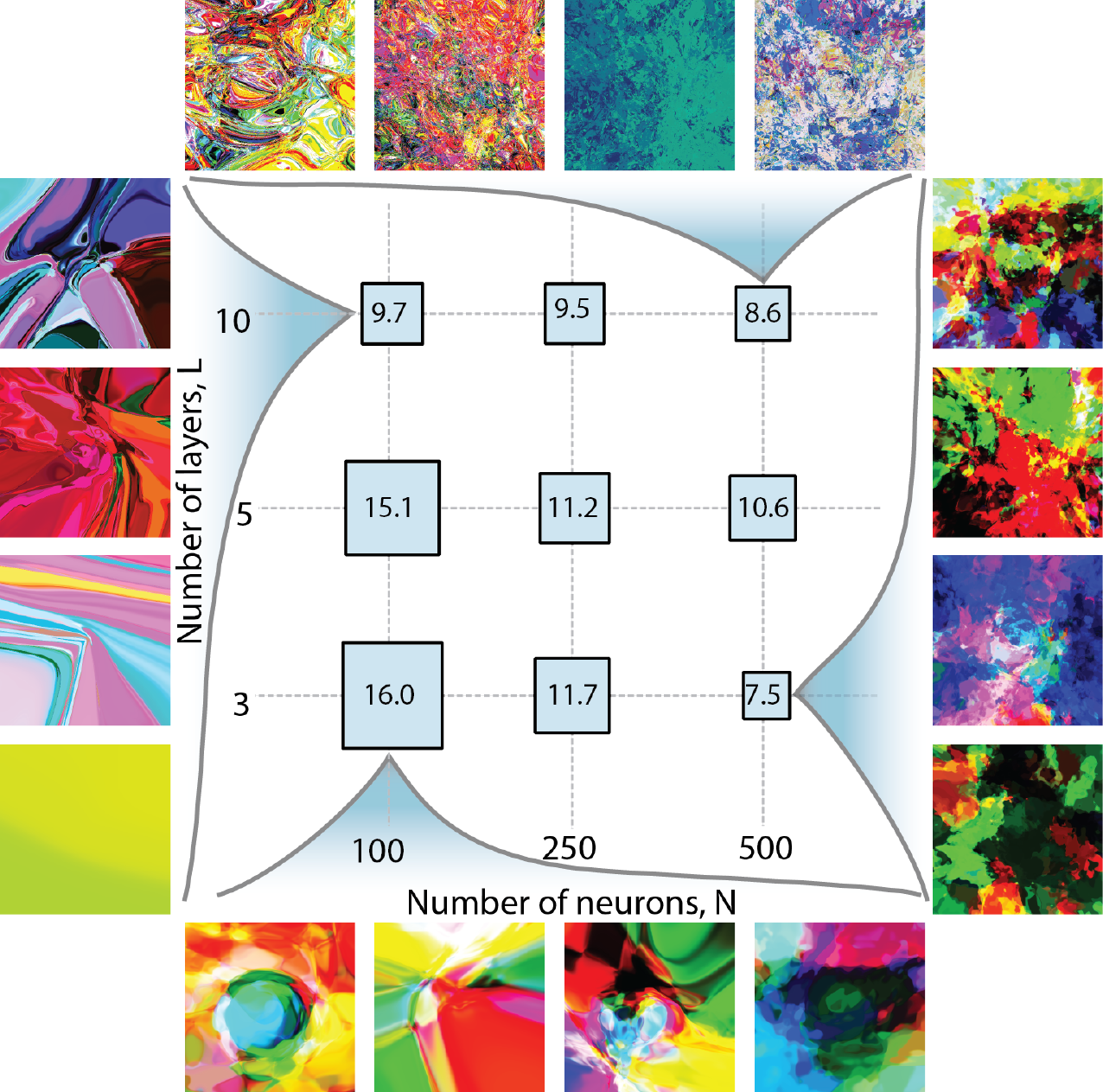}
    \caption{\textbf{Diversity of images generated by CPPNs and distribution of aesthetic images across architectures}.   All possible architectures are grouped according to their number of neurons and number of layers $(N, L)$ parameters.  Example images (not necessarily tagged as aesthetic) are shown for 4 of these classes.
    The sides of squares in the central panel are proportional to percentage of the aesthetically tagged images that belong to this class out of all coloured tagged images (the percentage is indicated in the middle of the square). The largest class, $(N, L) = (100, 3)$ contains 210 of 1314 tagged images.
    }
    \label{fig:images}
\end{figure}

\subsection{Influence of network architecture on image statistics}
\begin{figure*}[ptb]
    \centering
    \includegraphics[width=.89\textwidth]{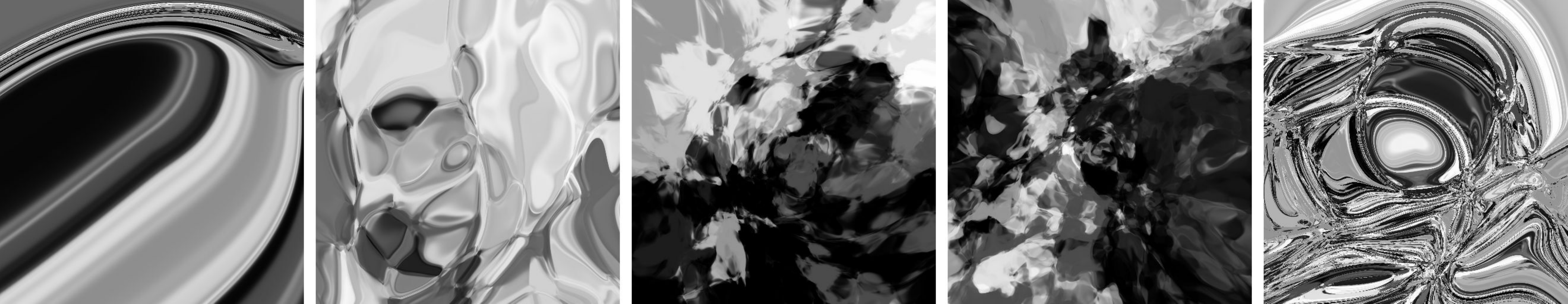}\\[4pt]
    \includegraphics[width=.89\textwidth]{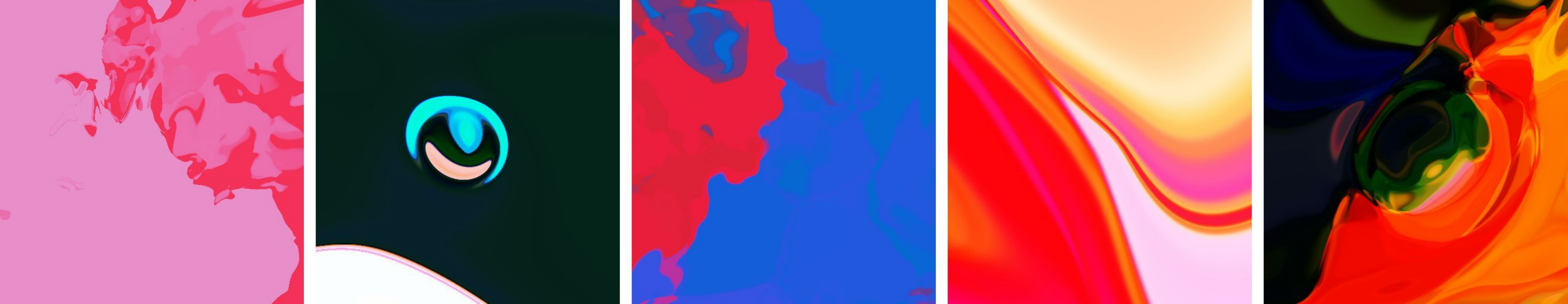}\\[4pt]
    \includegraphics[width=.89\textwidth]{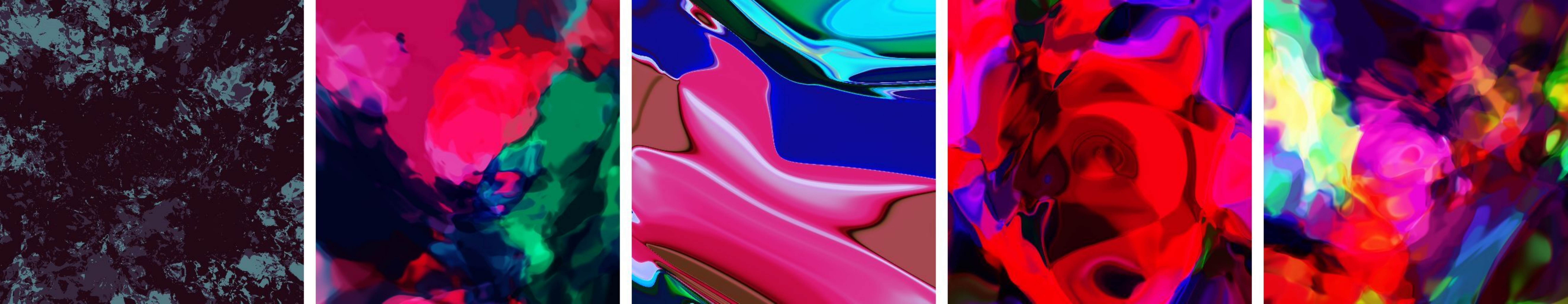}\\[4pt]
    \includegraphics[width=.89\textwidth]{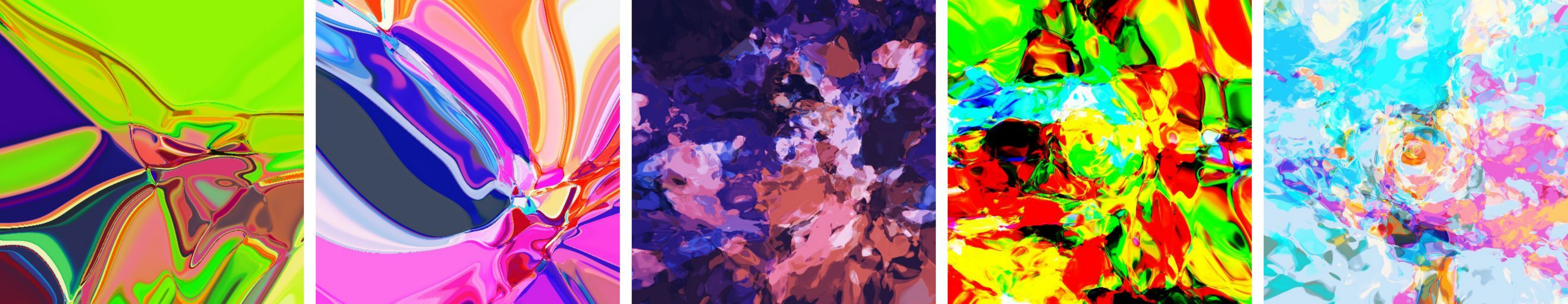}\\[4pt]
    \includegraphics[width=.89\textwidth]{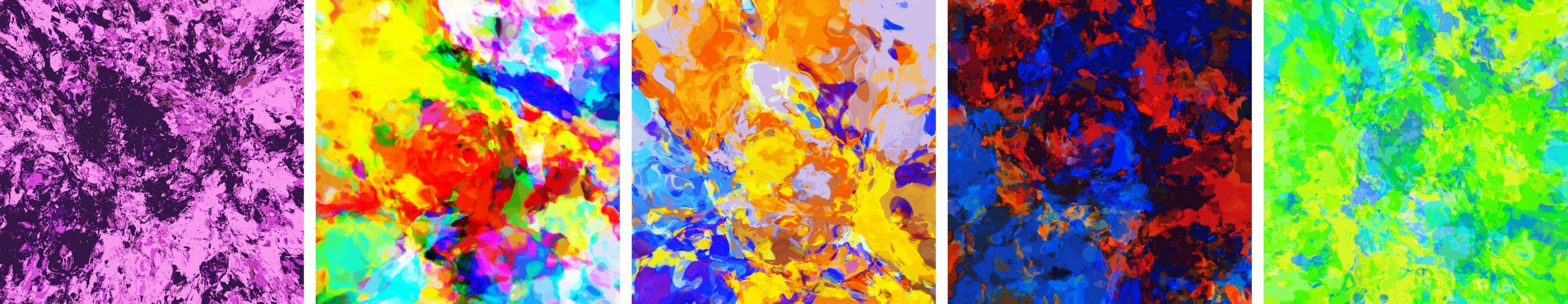}\\[4pt]
    \includegraphics[width=.89\textwidth]{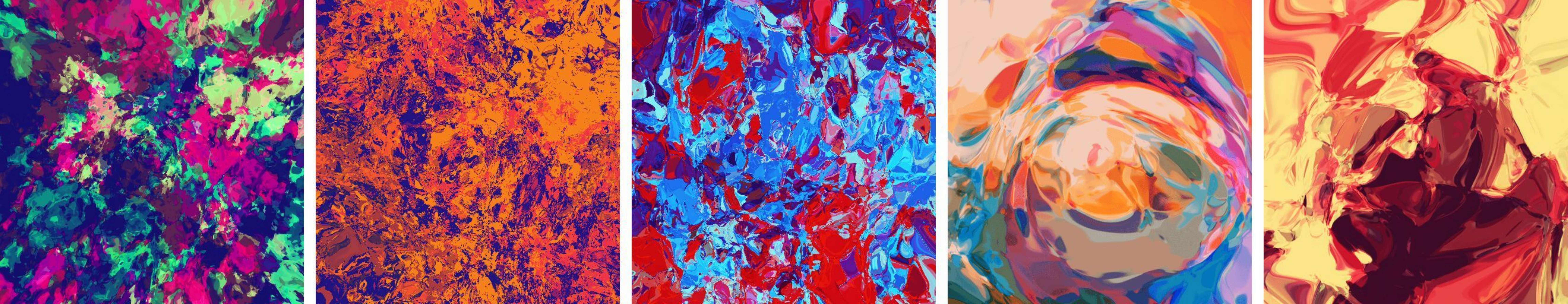}\\[4pt]
    \includegraphics[width=.89\textwidth]{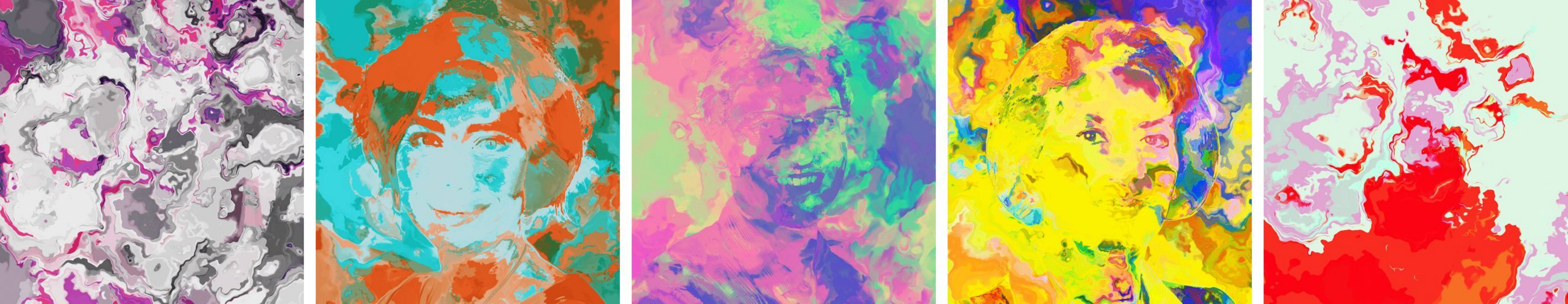}
    \caption{\textbf{Selection of aesthetically tagged CPPN generated images.}
      Top row shows grayscaled images.
      The last row is generated adding additional inputs (not used in the experiments).
      The two outer images are using 2D Perlin noise and the remaining three
      use pictures of faces as additional input.
     }
    \label{fig:many_pics}
\end{figure*}

Using the architecture parametrization equation (Eq.~\ref{eq:net_arch}) 882 unique architectures were defined and 40 images were generated per architecture for a total of 35280 images. Some architectures were found to produce images of similar quality and statistics while others were found to be comparatively unique in their outputs, see Fig.~\ref{fig:many_pics}.
Ultimately the parametrization prescribed in this paper allowed the CPPNs to generate a variety of patterns and forms (see Fig.~\ref{fig:images} for examples). Images with many layers tend to have fractal-like patterns with sharp edges and boundaries, as if the images from an architecture with less layers have folded in on themselves multiple times. Images with more neurons tend to have smaller scale patterns and clusters, and are generally noisier. However, the distribution of neurons across the layers is found to be of vital importance as well. For example if progressive layers in a network had a growing number of neurons, the resulting images it generates tends to have much smoother and simpler shapes than an architecture with the same number of layers and total neurons but with more neurons in its initial layers and decreasing neurons in progressively deeper layers.
The exponential growth/decay and periodic oscillation terms in the architecture parametrization Eq.~\ref{eq:net_arch} allow for the creation of a variety of patterns for the same number of layers and neurons.

\begin{figure}
    \centering
    \includegraphics[width=1\linewidth]{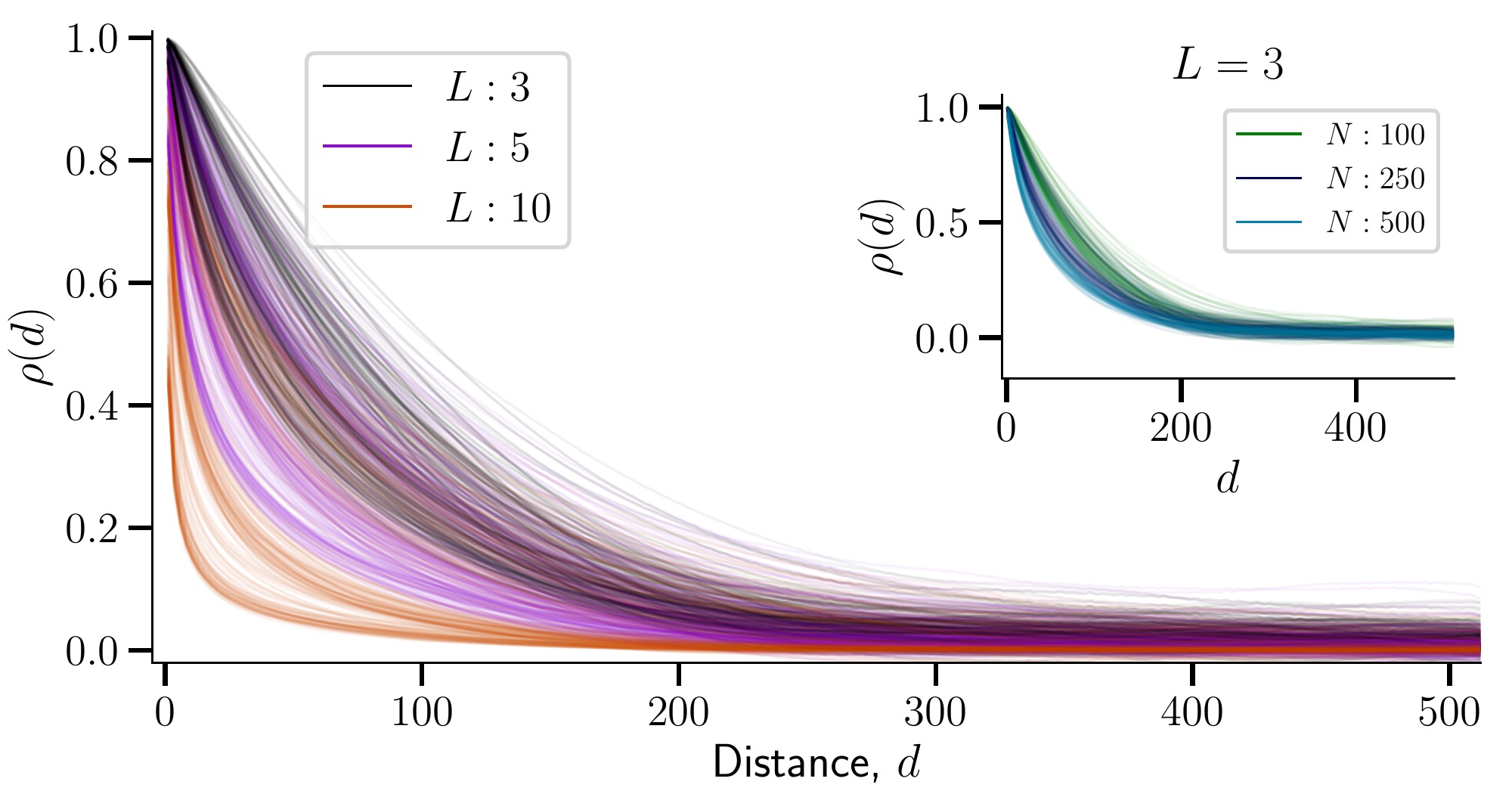}
    \caption{\textbf{Correlation functions of CPPN generated images averaged across architectures.} Each line corresponds to the mean correlation function of 40 images generated for each of the 882 architectures. The lines are then colour-coded according to the number of layers in that architecture. \textbf{Inset: }The correlation functions for all architectures with $L=3$ layers are plotted and colour-coded according to the number of neurons in the network.
    }
    \label{fig:correlation}
\end{figure}

The decay of spatial correlations can characterize the presence of particular scales or scale-freeness~\cite{cavagna2010scale}. We investigate the spatial correlation of all CPPN images. We average correlations of all 40 images of each network architecture and then plot all of the correlation functions together,  see Fig.~\ref{fig:correlation}. To understand how particular parameters used for the architecture generation impact the resulting correlations we colour-coded these correlation functions according to the number of layers ($L$) in their architectures.
The clustering of the curves into colour-bands indicates that the number of layers is a dictating parameter. Larger numbers of layers generally results in faster correlation decays. However, the curves begin to overlap  heavily in region between the fastest and slowest decaying curves.  The second most important parameter appeared to be the total number of neurons ($N$). For any given number of layers $L$, larger $N$ results in faster decaying correlation functions. For larger $L$ there is larger separation between groups of curves for different $N$. However, even for the smallest $L=3$ curves corresponding to the different $N$ are well-clustered as can be seen in the inset of Fig.~\ref{fig:correlation}. The other parameters in our parametrization, $(\alpha, \mu, \omega)$ also played an important role in the quality of the images produced, however, when analyzing the ensemble of image correlations, they were found to not be as important in characterizing their respective ensemble correlation functions.

\subsection{Features of aesthetic images}

\begin{figure}
    \centering
    \includegraphics[width=1\linewidth]{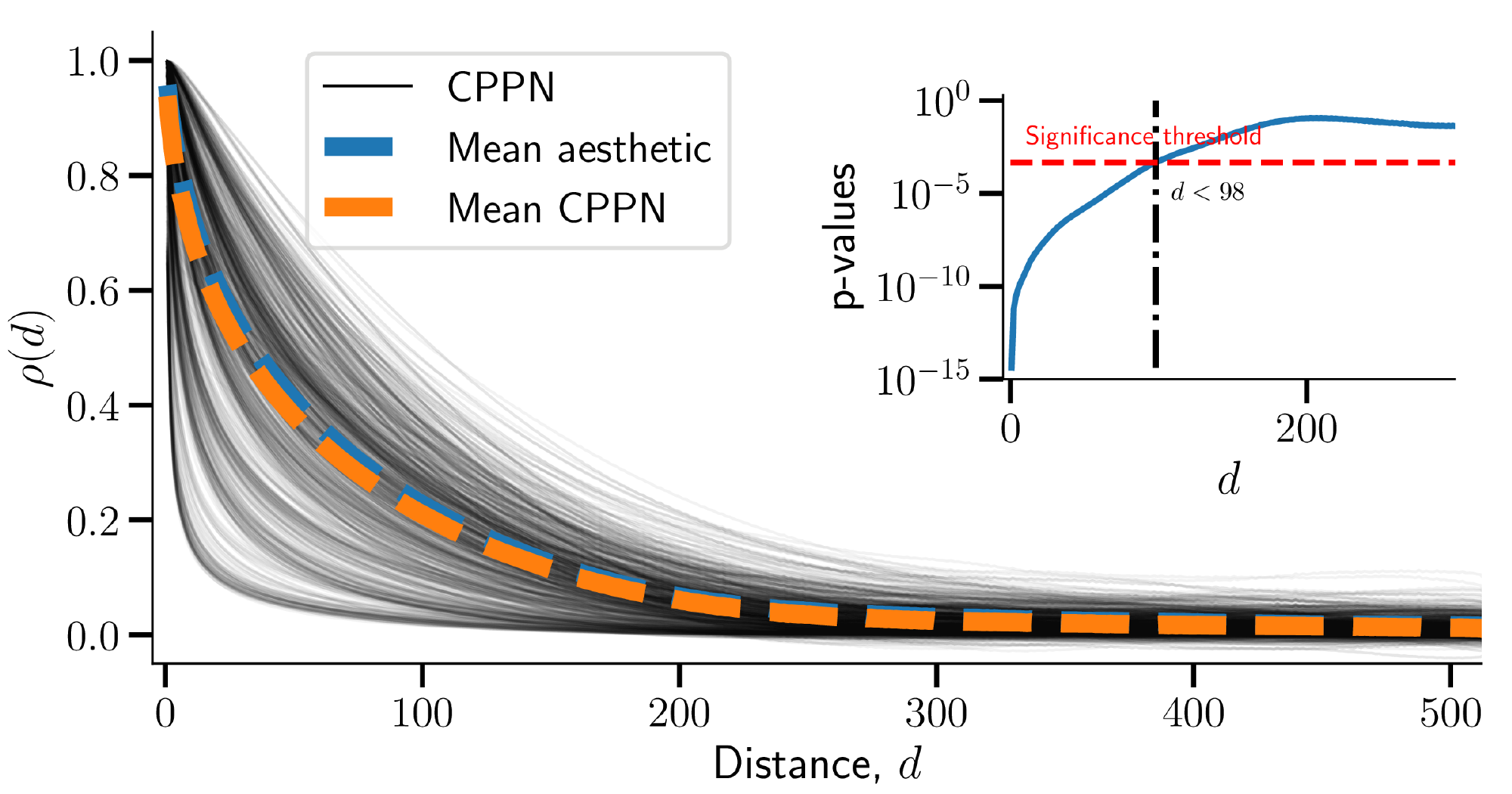}
    \caption{\textbf{Mean correlation functions of the full aesthetic image set and the full CPPN image set.} All CPPN architecture correlation functions (thin dark lines), the means of all CPPN correlation functions (thick dashed orange line), and the means of all the aesthetic images (thick blue dashed line) are plotted in the main figure.
\textbf{Inset: }The p-values for a one-sided Welch's test (for $d < 300$) between the means of the aesthetic set and the full CPPN set are plotted on a logarithmic scale.
The dashed line shows the Bonferonni-corrected threshold of significance at $p \ll 4\cdot10^{-4}$ revealing that the sets significantly differ for correlations $d < 98$.
    }
    \label{fig:p_global}
\end{figure}

We investigate the sets of 1403 (1314) grayscale (colour) images that were tagged as 'aesthetic' and compared it to the rest of the generated images.
Similar to a previous analysis, the spatial correlation functions of the positively tagged images were calculated.
We find that images with correlation functions that decay more slowly tended to be picked slightly more often.
We investigate this statement statistically using a one-sided Welch's test to compare the means of these sets of images against each other. A significant difference ($p \ll 4 \cdot 10^{-4}$ after Bonferroni correction for multiple comparisons) is found between the set of aesthetic images and the set of all CPPN images for distances up to 97 pixels as seen in the inset of Fig.~\ref{fig:p_global}.

\begin{figure}
    \centering
    \includegraphics[width=0.96\linewidth]{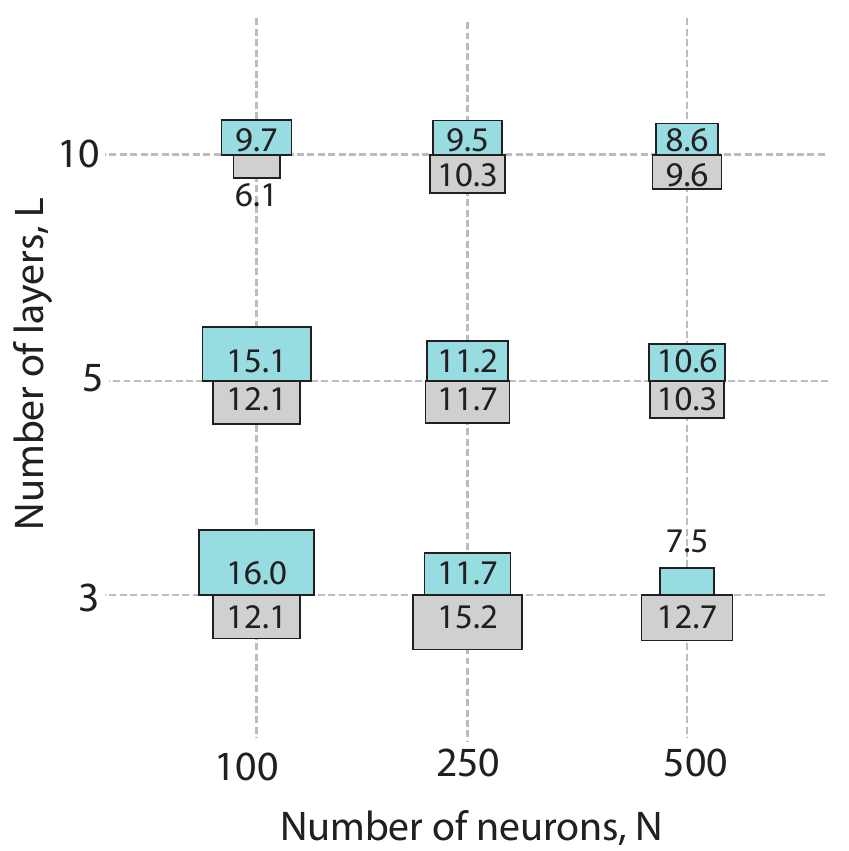}
    \caption{\textbf{Distribution of aesthetic grayscale and colour images  across different network  architectures}.  Grey boxes represent the grayscale images, blue boxes represent the colour images. The sides of the half-squares are proportional to the number of times an image of that class was tagged as aesthetic out of a total of 1403/1314 for grayscale/colour images respectively. The numbers inside the boxes represent the percentage of images tagged within that configuration. All possible architectures are grouped according to their number of neurons and number of layers $(N, L)$ parameters.
    }
    \label{fig:arch-color}
\end{figure}

In general, such significant difference might be explained if relatively simpler architectures that have a slower than average decay of correlation functions are selected more often.
Indeed, we observe in Fig.~\ref{fig:arch-color} that the architecture configurations selected most often among both the grayscale images and the colour images tended to trend toward fewer layers and fewer neurons, respectively.
These simpler architectures generally generated images with slower correlation decays than the architectures with more layers and neurons.
Notably, participants were more willing to tag images from more complex architectures with more neurons when judging grayscale images than in colour images.
Participants reported that the colour images felt more chaotic than the grayscale and would opt for simpler, smoother structures.
Interestingly, while the distribution of aesthetic architecture configurations varied somewhat dramatically between the colour set and the grayscale set, mean correlation functions between these two aesthetic image sets did not show significant differences. This implies that the qualitative differences between these architectures may not show up when comparing their mean correlation functions.

To check if the slower-decaying trends observed globally among the aesthetic images were also observable \textit{within} architectures, we compared the aesthetic images of a particular $(N, L)$ architecture class with the larger population of images within that class. In the same way as before, we apply a Welch's test to compare the correlation functions between the sets and recorded a Bonferroni corrected p-value. By permuting all combinations of $N$ and $L$, 9 such comparisons can be made for both colour and grayscale image sets. Overall, no obviously significant trend was found between these sub-sets of aesthetic images and their respective architecture classes. This implies that the global behaviour of the entire set of aesthetic images is most strongly affected by the bias towards simpler architectures than any potential exceptionalism of images within their architectures.

\subsection{Comparison between generated images and natural images}\label{sec:comp:gen-nat}

\begin{figure*}
    \centering
    \includegraphics[width=1\textwidth]{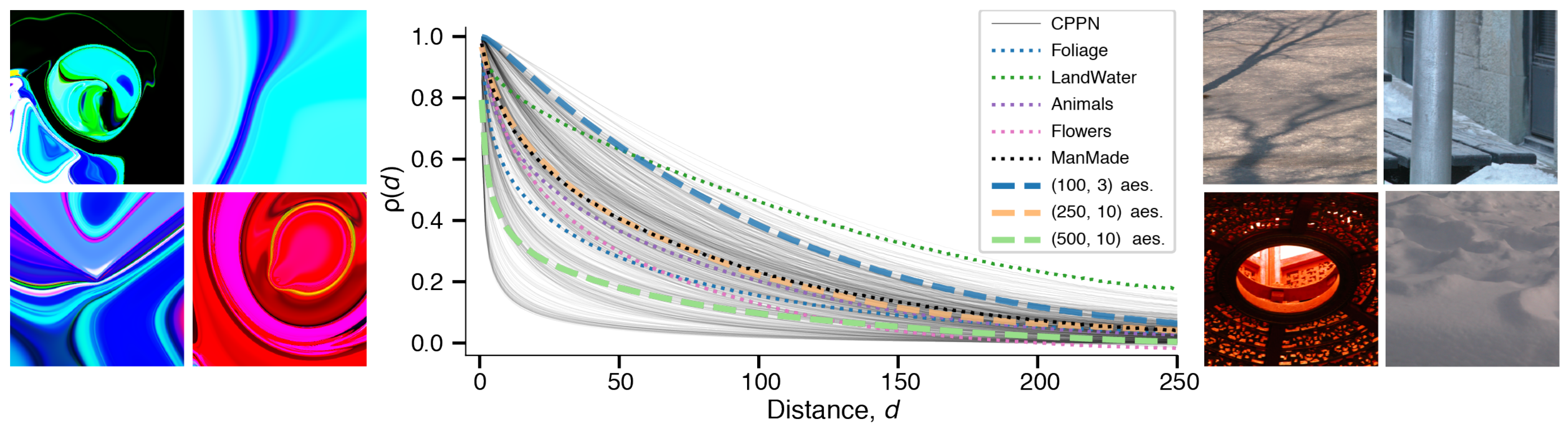}
    \caption{\textbf{Comparison of natural images and CPPN generated images.} Central panel: decay of the two-point correlation as function of distance in different classes of natural images (see legend) and in the groups of aesthetic images corresponding to a particular $(N,L)$ pairs. Left panel: examples of tagged aesthetic images from the $(250,10)$ group (that corresponds to an orange line in the central panel). Right panel: examples of natural images from `ManMade' group (corresponds to the black-dotted line in the central panel) and `Snow' group (omitted for clarity due to overlapping with the `ManMade' curve).
     }
    \label{fig:natural_vs_aes}
\end{figure*}
Natural images statistics is influenced by many different parameters, including zoom-levels, scene composition, and so forth~\cite{wainwright2000scale,hyvarinen2009natural, gerhard2015modeling}.  Consequently, natural images also posses a broad range of different correlation functions.
To test how well the CPPN architectures parametrization could generate a broad space of correlation functions and image structures, we compare its statistics with different sets of natural images, see Fig.~\ref{fig:natural_vs_aes}. Unsurprisingly, different categories of natural images produce a variety of correlation functions. For example the category `LandWater' have smaller correlations at shorter distances and larger correlations at longer distances and is not particularly well represented by any of the CPPN architectures although images with similar statistics are generated.
Meanwhile the categories of `Foliage', `Flowers', and `Animals' have faster decaying correlation functions and were comparable to the slightly noisier CPPN architectures with 5 or 10 layers. In general however, our parametrization seems capable of generating images from a broad space of statistics, at least when compared to these image sets.

A common perception reported by participants was the sense of aesthetics in abstractions arising from the interpretation and identification of familiar objects. Perhaps the aesthetic CPPN images can elicit reactions that resemble those to natural stimuli due to their statistical similarity. To explore in this direction, natural image statistics are also compared to the aesthetic image sets which are categorized according to their $(N,L)$ architecture parameters. These correlation functions are shown on top of the full set of CPPN correlation functions in Fig.~\ref{fig:natural_vs_aes}. It can be seen in this figure that across the bandwidth of correlation functions that are generated by our CPPN architectures, that the characteristic correlation functions of the natural image datasets can be represented by some architectures in our parametrization. The similar correlation functions of the $(N, L) = (250, 10)$ CPPN architecture are compared to the `ManMade' and `Snow' natural image set with some example images from each set (the `Snow' correlation function overlapped almost identically with the `ManMade' and was omitted for visual clarity). Though the correlation functions of these sets are very similar, it is easy to see that individual images are still very much different. However, some similarities in the structure and distribution of patterns and shapes in the sets can still be observed (see the panels on the side of Fig.~\ref{fig:natural_vs_aes}).

\subsection{Limitations}
To extract an elusive `aesthetic' signal from the noise and variety of human perception and subjectivity is a difficult task. Controlling for human variability and subjectivity demands that we have an understanding of the scope and scale of human variability in this abstract perceptual space, something we do not have. With only 45 subjects, a notion of perceptual clusters began to emerge where some distinct groups of individuals with somewhat similar tastes. However, we still could not look at this sparsely populated space as an ensemble in the same way we could within our neural network architecture space. Gathering data from social media where millions of people around the world voice their opinion is one method that has been employed in the past to alleviate this problem \cite{schwarz2018will}. However, tracking an individual's perceptual space can still be difficult online due to its potentially anonymous nature.

\section{Conclusion and Discussion}
In the present work, we explore the capacity of randomly generated CPPNs to generate aesthetic abstract imagery and quantify them using their two-point auto-correlation.
We analyse which network architectures with random weights likely generate such statistics. By comparing images that were deemed aesthetic by our volunteer participants we find that in general aesthetic images tend to have correlation functions that decay slightly slower. This result holds true globally but not locally, within individual architecture classes.

It is worth mentioning that any deviations from the mean correlation functions of the full dataset is highly dependent on the parametrization we employ in our architecture sampling equation. If the set of architectures are equally balanced with `simple' and `complex' images in perceptual space, then averaging over the correlations of aesthetic images may yield the same global average across all images.
In contrast, had we selected more complex architectures which produce noisier images, we might observe that the mean correlations of the aesthetic images would be drastically different from those of the full dataset.

When comparing the generated images to natural images we find that our architecture parametrization is capable of representing statistics similar to a variety of natural images.

Some participants reported seeing objects, animals, cars or faces in some images.

Finding familiarity in such abstractions can itself be a source of aesthetic appeal, at least for a period of time until the novelty wears off~\cite{hertzmann2019aesthetics, schmidhuber2010formal}.
Generally speaking, the battle between order and complexity, `unity in variety', is an often discussed topic seemingly at the heart of the emergence of the perception of aesthetics \cite{neumann2005defining}.

This study does not explicitly check for power-laws in the Fourier spectrum of the images (which are commonly associated with natural image statistics~\cite{ruderman1994statistics} and are related to the correlation function under some assumptions by the Wiener-Khinchin theorem). We observe, however, in the spatial domain of correlations that aesthetic images tend to have lower decay rates than other CPPN generated images. This long tail of correlations may be rooted in self-similarity which has been observed to be strongly correlated with high aesthetic scores in experiments analyzing the aesthetic scores of artistic imagery~\cite{amirshahi2012phog, amirshahi2013self, melmer2013regular, denzler2016convolutional}.

Earlier, multiple measures to evaluate aesthetic features of photographs and images were studied in the contexts of automatic image selection~\cite{deng2017image, datta2006studying, wang2018neural}, or evolutionary generation of aesthetic images~\cite{den2010using}.
Human subjects were shown to be very sensitive to the statistics of local textures~\cite{gerhard2014towards}.
Here we demonstrate that humans prefer images with large correlations at scales exceeding small textured segments.

It can be understood how CPPNs are able to generate such smooth and structured imagery by virtue of the correlation between the inputs that flow independently through the random network.
Even though the weights of the network are generated randomly, the inherent correlations in the input can still survive.
Smooth/coherent noise has long been used in computer graphics to generate realistic textures and visuals, for example in Perlin noise~\cite{perlin1985image, lagae2010survey}.
Interestingly, CPPNs seem to reverse the order of operations. Instead of starting from a noisy grid and smoothing out the points in-between, CPPNs start with smooth images and progressively add randomness (by random projections and non-linearities) layer by layer.

Thus, the underlying transformation is not overly complicated, transforming boring smoothness into structured complexity. By creating similar statistical features as natural images it creates
 familiar statistics, order, and patterns from pseudo-randomness which humans tend to find delightful~\cite{schmidhuber2010formal}.

Even though the randomness in the CPPN weights suggests that one has little control over the output of the network, in reality the architecture narrows the type of the picture quite strongly.

For example having the final layer with only 1 neuron ensures monochrome images; 3 neurons can be red, green, blue; 4 neurons can be red, green blue, alpha; having a bottleneck layer of only a few neurons in the middle of the architecture can act to simplify the colour palette of the final image; having many layers with a small number of neurons can generate fractal-like images.
A variety of interesting architectures capable of generating extremely unique image styles were not included in this project as we opted to study a relatively simple architecture parametrization to keep our variables tractable.
Many studies related to weight-agnostic neural networks have found the predisposed biases inherent to particular architectures allowing some of them to perform well without training~\cite{gaier2019weight, ulyanov2018deep}.

We believe CPPNs are an excellent source of unbiased abstract images that may help our understanding on how we define aesthetics and its mathematical boundaries.

\section*{Acknowledgment}

The authors would like to thank David Ha (@hardmaru) for his insightful tutorial on CPPNs.
AL received funding from a Sofja Kovalevskaja Award
from the Alexander von Humboldt Foundation, endowed by the Federal Ministry of
Education and Research. We acknowledge support from the BMBF through the T\"ubingen AI Center (FKZ: 01IS18039B).

\printbibliography

@article{ruderman1994statistics,
	title={The statistics of natural images},
	author={Ruderman, Daniel L},
	journal={Network: computation in neural systems},
	volume={5},
	number={4},
	pages={517--548},
	year={1994},
	publisher={Taylor \& Francis}
}

@article{olshausen1996emergence,
  title={Emergence of simple-cell receptive field properties by learning a sparse code for natural images},
  author={Olshausen, Bruno A and Field, David J},
  journal={Nature},
  volume={381},
  number={6583},
  pages={607},
  year={1996},
  publisher={Nature Publishing Group}
}

@incollection{gerhard2015modeling,
  title={Modeling natural image statistics},
  author={Gerhard, Holly E and Theis, Lucas and Bethge, Matthias},
  booktitle={Biologically Inspired Computer Vision},
  pages={53--80},
  year={2015},
  publisher={Wiley Online Library}
}

@book{hyvarinen2009natural,
  title={Natural image statistics: A probabilistic approach to early computational vision.},
  author={Hyv{\"a}rinen, Aapo and Hurri, Jarmo and Hoyer, Patrick O},
  volume={39},
  year={2009},
  publisher={Springer Science \& Business Media}
}

@inproceedings{wainwright2000scale,
  title={Scale mixtures of Gaussians and the statistics of natural images},
  author={Wainwright, Martin J and Simoncelli, Eero P},
  booktitle={Advances in neural information processing systems},
  pages={855--861},
  year={2000}
}

@article{olmos2004biologically,
  title={A biologically inspired algorithm for the recovery of shading and reflectance images},
  author={Olmos, Adriana and Kingdom, Frederick AA},
  journal={Perception},
  volume={33},
  number={12},
  pages={1463--1473},
  year={2004},
  publisher={SAGE Publications Sage UK: London, England}
}

@article{deng2017image,
  title={Image aesthetic assessment: An experimental survey},
  author={Deng, Yubin and Loy, Chen Change and Tang, Xiaoou},
  journal={IEEE Signal Processing Magazine},
  volume={34},
  number={4},
  pages={80--106},
  year={2017},
  publisher={IEEE}
}

@inproceedings{den2010using,
  title={Using aesthetic measures to evolve art},
  author={den Heijer, Eelco and Eiben, AE},
  booktitle={IEEE Congress on Evolutionary Computation},
  pages={1--8},
  year={2010},
  organization={IEEE}
}

@inproceedings{datta2006studying,
  title={Studying aesthetics in photographic images using a computational approach},
  author={Datta, Ritendra and Joshi, Dhiraj and Li, Jia and Wang, James Z},
  booktitle={European conference on computer vision},
  pages={288--301},
  year={2006},
  organization={Springer}
}

@inproceedings{nguyen2015deep,
  title={Deep neural networks are easily fooled: High confidence predictions for unrecognizable images},
  author={Nguyen, Anh and Yosinski, Jason and Clune, Jeff},
  booktitle={Proceedings of the IEEE conference on computer vision and pattern recognition},
  pages={427--436},
  year={2015}
}

@article{gerhard2014towards,
  title={Towards rigorous study of artistic style: a new psychophysical paradigm},
  author={Gerhard, Holly E and Bethge, Matthias},
  journal={Art \& Perception},
  volume={2},
  number={1-2},
  pages={23--44},
  year={2014},
  publisher={Brill}
}

@article{stanley2007compositional,
  title={Compositional pattern producing networks: A novel abstraction of development},
  author={Stanley, Kenneth O},
  journal={Genetic programming and evolvable machines},
  volume={8},
  number={2},
  pages={131--162},
  year={2007},
  publisher={Springer}
}

@inproceedings{secretan2008picbreeder,
  title={Picbreeder: evolving pictures collaboratively online},
  author={Secretan, Jimmy and Beato, Nicholas and D Ambrosio, David B and Rodriguez, Adelein and Campbell, Adam and Stanley, Kenneth O},
  booktitle={Proceedings of Computer Human Interaction Conference (CHI 2008)},
  pages={1759--1768},
  year={2008},
  organization={ACM}
}

@article{Kingma2013AutoEncodingVB,
  title={Auto-Encoding Variational Bayes},
  author={Diederik P. Kingma and Max Welling},
  journal={CoRR},
  year={2013},
  volume={abs/1312.6114}
}

@incollection{Goodfellow2014:GANS,
title = {Generative Adversarial Nets},
author = {Goodfellow, Ian and Pouget-Abadie, Jean and Mirza, Mehdi and Xu, Bing and Warde-Farley, David and Ozair, Sherjil and Courville, Aaron and Bengio, Yoshua},
booktitle = {Advances in Neural Information Processing Systems 27},
editor = {Z. Ghahramani and M. Welling and C. Cortes and N. D. Lawrence and K. Q. Weinberger},
pages = {2672--2680},
year = {2014},
publisher = {Curran Associates, Inc.},
url = {http://papers.nips.cc/paper/5423-generative-adversarial-nets.pdf}
}

@article{gaier2019weight,
  title={Weight Agnostic Neural Networks},
  author={Gaier, Adam and Ha, David},
  journal={arXiv preprint arXiv:1906.04358},
  year={2019}
}

@inproceedings{ulyanov2018deep,
  title={Deep image prior},
  author={Ulyanov, Dmitry and Vedaldi, Andrea and Lempitsky, Victor},
  booktitle={Proceedings of the IEEE Conference on Computer Vision and Pattern Recognition},
  pages={9446--9454},
  year={2018}
}

@article{perlin1985image,
  title={An image synthesizer},
  author={Perlin, Ken},
  journal={ACM Siggraph Computer Graphics},
  volume={19},
  number={3},
  pages={287--296},
  year={1985}
}

@inproceedings{lagae2010survey,
  title={A survey of procedural noise functions},
  author={Lagae, Ares and Lefebvre, Sylvain and Cook, Rob and DeRose, Tony and Drettakis, George and Ebert, David S and Lewis, John P and Perlin, Ken and Zwicker, Matthias},
  booktitle={Computer Graphics Forum},
  volume={29},
  number={8},
  pages={2579--2600},
  year={2010},
  organization={Wiley Online Library}
}

@article{schmidhuber2010formal,
  title={Formal theory of creativity, fun, and intrinsic motivation (1990--2010)},
  author={Schmidhuber, J{\"u}rgen},
  journal={IEEE Transactions on Autonomous Mental Development},
  volume={2},
  number={3},
  pages={230--247},
  year={2010},
  publisher={IEEE}
}

@article{wang2018neural,
  title={Neural Aesthetic Image Reviewer},
  author={Wang, Wenshan and Yang, Su and Zhang, Weishan and Zhang, Jiulong},
  journal={arXiv preprint arXiv:1802.10240},
  year={2018}
}

@inproceedings{amirshahi2012phog,
  title={PHOG analysis of self-similarity in aesthetic images},
  author={Amirshahi, Seyed Ali and Koch, Michael and Denzler, Joachim and Redies, Christoph},
  booktitle={Human Vision and Electronic Imaging XVII},
  volume={8291},
  pages={82911J},
  year={2012},
  organization={International Society for Optics and Photonics}
}

@inproceedings{amirshahi2013self,
  title={How self-similar are artworks at different levels of spatial resolution?},
  author={Amirshahi, Seyed Ali and Redies, Christoph and Denzler, Joachim},
  booktitle={Proceedings of the Symposium on Computational Aesthetics},
  pages={93--100},
  year={2013},
  organization={ACM}
}

@article{melmer2013regular,
  title={From regular text to artistic writing and artworks: Fourier statistics of images with low and high aesthetic appeal},
  author={Melmer, Tamara and Amirshahi, Seyed Ali and Koch, Michael and Denzler, Joachim and Redies, Christoph},
  journal={Frontiers in human neuroscience},
  volume={7},
  pages={106},
  year={2013},
  publisher={Frontiers}
}

@inproceedings{denzler2016convolutional,
  title={Convolutional neural networks as a computational model for the underlying processes of aesthetics perception},
  author={Denzler, Joachim and Rodner, Erik and Simon, Marcel},
  booktitle={European Conference on Computer Vision},
  pages={871--887},
  year={2016},
  organization={Springer}
}

@article{hertzmann2019aesthetics,
  title={Aesthetics of Neural Network Art},
  author={Hertzmann, Aaron},
  journal={arXiv preprint arXiv:1903.05696},
  year={2019}
}

@article{cavagna2010scale,
  title={Scale-free correlations in starling flocks},
  author={Cavagna, Andrea and Cimarelli, Alessio and Giardina, Irene and Parisi, Giorgio and Santagati, Raffaele and Stefanini, Fabio and Viale, Massimiliano},
  journal={Proceedings of the National Academy of Sciences},
  volume={107},
  number={26},
  pages={11865--11870},
  year={2010},
  publisher={National Acad Sciences}
}

@inproceedings{schwarz2018will,
  title={Will people like your image? learning the aesthetic space},
  author={Schwarz, Katharina and Wieschollek, Patrick and Lensch, Hendrik PA},
  booktitle={2018 IEEE Winter Conference on Applications of Computer Vision (WACV)},
  pages={2048--2057},
  year={2018},
  organization={IEEE}
}

@article{neumann2005defining,
  title={Defining computational aesthetics},
  author={Neumann, L and Sbert, M and Gooch, B and Purgathofer, W and others},
  journal={Computational aesthetics in graphics, visualization and imaging},
  pages={13--18},
  year={2005},
  publisher={Citeseer}
}

@Misc{XnView,
  author =    {Gougelet Pierre-Emmanuel},
  title =     {XnView Image viewer},
  howpublished = {\url{www.xnview.com}},
  year =      2019}

\end{document}